\documentclass[screen,sigconf]{acmart}

\AtBeginDocument{%
  \providecommand\BibTeX{{%
    \normalfont B\kern-0.5em{\scshape i\kern-0.25em b}\kern-0.8em\TeX}}}

\acmYear{2021}\copyrightyear{2021}
\setcopyright{rightsretained}
\acmConference[ACM CHIL '21]{ACM Conference on Health, Inference, and Learning}{April 8--10, 2021}{Virtual Event, }
\acmBooktitle{ACM Conference on Health, Inference, and Learning (ACM CHIL '21), April 8--10, 2021, Virtual Event, }
\acmPrice{}
\acmDOI{10.1145/3450439.3451867}
\acmISBN{978-1-4503-8359-2/21/04}

\begin{document}

\title{CheXtransfer: Performance and Parameter Efficiency of ImageNet Models for Chest X-Ray Interpretation}


\author{Alexander Ke}
\authornote{Authors contributed equally to this research.}
\email{alexke@cs.stanford.edu}
\affiliation{
    \institution{Stanford University}
    \country{USA}
}
\author{William Ellsworth}
\authornotemark[1]
\email{willells@cs.stanford.edu}
\affiliation{
    \institution{Stanford University}
    \country{USA}
}
\author{Oishi Banerjee}
\authornotemark[1]
\email{oishi.banerjee@cs.stanford.edu}
\affiliation{
    \institution{Stanford University}
    \country{USA}
}
\author{Andrew Y. Ng}
\email{ang@cs.stanford.edu}
\affiliation{
    \institution{Stanford University}
    \country{USA}
}
\author{Pranav Rajpurkar}
\email{pranavsr@cs.stanford.edu}
\affiliation{
    \institution{Stanford University}
    \country{USA}
}

\begin{abstract}
Deep learning methods for chest X-ray interpretation typically rely on pretrained models developed for ImageNet. This paradigm assumes that better ImageNet architectures perform better on chest X-ray tasks and that ImageNet-pretrained weights provide a performance boost over random initialization. In this work, we compare the transfer performance and parameter efficiency of 16 popular convolutional architectures on a large chest X-ray dataset (CheXpert) to investigate these assumptions. First, we find no relationship between ImageNet performance and CheXpert performance for both models without pretraining and models with pretraining. Second, we find that, for models without pretraining, the choice of model family influences performance more than size within a family for medical imaging tasks. Third, we observe that ImageNet pretraining yields a statistically significant boost in performance across architectures, with a higher boost for smaller architectures. Fourth, we examine whether ImageNet architectures are unnecessarily large for CheXpert by truncating final blocks from pretrained models, and find that we can make models 3.25x more parameter-efficient on average without a statistically significant drop in performance. Our work contributes new experimental evidence about the relation of ImageNet to chest x-ray interpretation performance.
\end{abstract}



\begin{teaserfigure}
  \includegraphics[width=\textwidth]{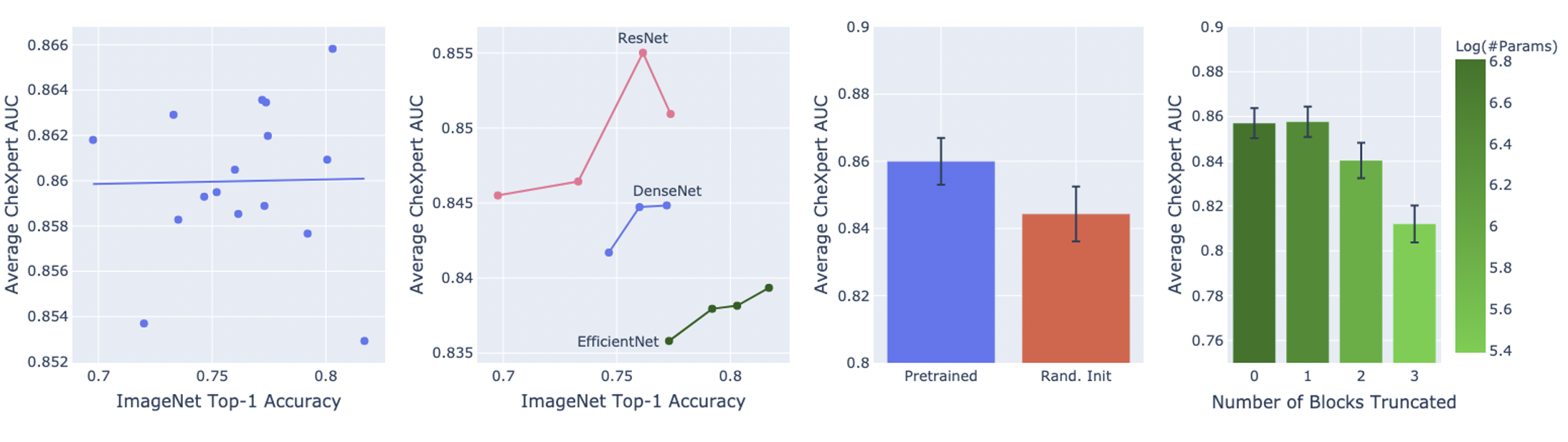}
  \caption{Visual summary of our contributions. From left to right: scatterplot and best-fit line for 16 pretrained models showing no relationship between ImageNet and CheXpert performance, CheXpert performance relationship varies across architecture families much more than within, average CheXpert performance improves with pretraining, models can maintain performance and improve parameter efficiency through truncation of final blocks. Error bars show one standard deviation.}
  \Description{See Introduction for contributions.}
  \label{fig:teaser}
\end{teaserfigure}

\maketitle

\section{Introduction}

Deep learning models for chest X-ray interpretation have high potential for social impact by aiding clinicians in their workflow and increasing access to radiology expertise worldwide \cite{rajpurkar2020chexaid, rajpurkar2021chexternal}. Transfer learning using pretrained ImageNet \cite{ImageNet} models has been the standard approach for developing models not only on chest X-rays \cite{chest1,chest2,covid} but also for many other medical imaging modalities \cite{Mitani2020,zhang2020retina,li2019retina,retina3,skincancer}. This transfer assumes that better ImageNet architectures perform better and pretrained weights boost performance on their target medical tasks. However, there has not been a systematic investigation of how ImageNet architectures and weights both relate to performance on downstream medical tasks.



In this work, we systematically investigate how ImageNet architectures and weights both relate to performance on chest X-ray tasks. Our primary contributions are:



\begin{enumerate}
    \item For models without pretraining and models with pretraining, we find \textit{no relationship between ImageNet performance and CheXpert performance} (Spearman $\rho = 0.08$, $\rho = 0.06$ respectively). This finding suggests that architecture improvements on ImageNet may not lead to improvements on medical imaging tasks.
    
    \item For models without pretraining, we find that within an architecture family, the largest and smallest models have small differences (ResNet 0.005, DenseNet 0.003, EfficientNet 0.004) in CheXpert AUC, but different model families have larger differences in AUC ($>0.006$). This finding suggests that \textit{the choice of model family influences performance more than size within a family} for medical imaging tasks.
    \item We observe that \textit{ImageNet pretraining yields a statistically significant boost in performance} (average boost of 0.016 AUC) across architectures, with a higher boost for smaller architectures (Spearman $\rho=-0.72$ with number of parameters). This finding supports the ImageNet pretraining paradigm for medical imaging tasks, especially for smaller models.
    \item We find that by truncating final blocks of pretrained models, we can make models \textit{3.25x more parameter-efficient on average without a statistically significant drop in performance}. This finding suggests model truncation may be a simple method to yield lighter pretrained models by preserving architecture design features while reducing model size.
\end{enumerate}

Our study, to the best of our knowledge, contributes the first systematic investigation of the performance and efficiency of ImageNet architectures and weights for chest X-ray interpretation. Our investigation and findings may be further validated on other datasets and medical imaging tasks.


\section{Related Work}

\subsection{ImageNet Transfer}

\citet{imagenettransfer} examined the performance of 16 convolutional neural networks (CNNs) on 12 image classification datasets. They found that using these ImageNet pretrained architectures either as feature extractors for logistic regression or fine tuning them on the target dataset yielded a Spearman $\rho = 0.99$ and $\rho = 0.97$ between ImageNet accuracy and transfer accuracy respectively. However, they showed ImageNet performance was less correlated with transfer accuracy for some fine-grained tasks, corroborating \citet{pretraining}. They found that without ImageNet pretraining, ImageNet accuracy and transfer accuracy had a weaker Spearman $\rho = 0.59$. We extend \citet{imagenettransfer} to the medical setting by studying the relationship between ImageNet and CheXpert performance.

\citet{transfusion} explored properties of transfer learning onto retinal fundus images and chest X-rays. They studied ResNet50 and InceptionV3 and showed pretraining offers little performance improvement. Architectures composed of just four to five sequential convolution and pooling layers achieved comparable performance on these tasks as ResNet50 with less than 40\% the parameters. In our work, we find pretraining does not boost performance for ResNet50, InceptionV3, InceptionV4, and MNASNet but does boost performance for the remaining 12 architectures. Thus, we were able to replicate \citet{transfusion}'s results, but upon studying a broader set of newer and more popular models, we reached the opposite conclusion that ImageNet pretraining yields a statistically significant boost in performance.

\subsection{Medical Task Architectures}

\citet{CheXpert} compared the performance of ResNet152, DenseNet121, InceptionV4, and SEResNeXt101 on CheXpert, finding that DenseNet121 performed best. In a recent analysis, all but one of the top ten CheXpert competition models used DenseNets as part of their ensemble, even though they have been surpassed on ImageNet \cite{rajpurkar2020chexpedition}. Few groups design their own networks from scratch, preferring to use established ResNet and DenseNet architectures for CheXpert \cite{bressem2020comparing}. This trend extends to retinal fundus and skin cancer tasks as well, where Inception architectures remain popular \cite{Mitani2020,zhang2020retina,li2019retina,retina3}. The popularity of these older ImageNet architectures hints that there may be a disconnect between ImageNet performance and medical task performance for newer architectures generated through architecture search. We verify that these newer architectures generated through search (EfficientNet, MobileNet, MNASNet) underperform older architectures (DenseNet, ResNet) on CheXpert, suggesting that search has overfit to ImageNet and explaining the popularity of these older architectures in the medical imaging literature.

\citet{bressem2020comparing} postulated that deep CNNs that can represent more complex relationships for ImageNet may not be necessary for CheXpert, which has greyscale inputs and fewer image classes. They studied ResNet, DenseNet, VGG, SqueezeNet, and AlexNet performance on CheXpert and found that ResNet152, DenseNet161, and ResNet50 performed best on CheXpert AUC. In terms of AUPRC, they showed that smaller architectures like AlexNet and VGG can perform similarly to deeper architectures on CheXpert. Models such as AlexNet, VGG, and SqueezeNet are no longer popular in the medical setting, so in our work, we systematically investigate the performance and efficiency of 16 more contemporary ImageNet architectures with and without pretraining. Additionally, we extend \cite{bressem2020comparing} by studying the effects of pretraining, characterizing the relationship between ImageNet and CheXpert performance, and drawing conclusions about architecture design.

\subsection{Truncated Architectures}


The more complex a convolutional architecture becomes, the more computational and memory resources are needed for its training and deployment. Model complexity thus may impede the deployment of CNNs to clinical settings with less resources. Therefore, efficiency, often reported in terms of the number of parameters in a model, the number of FLOPS in the forward pass, or the latency of the forward pass, has become increasingly important in model design. Low-rank factorization \cite{lowrank,xception}, transferred/compact convolutional filters \cite{compact}, knowledge distillation \cite{hinton2015distilling}, and parameter pruning \cite{parameter_pruning} have all been proposed to make CNNs more efficient.

Layer-wise pruning is a type of parameter pruning that locates and removes layers that are not as useful to the target task \cite{ro2020layerwise}. Through feature diagnosis, a linear classifier is trained using the feature maps at intermediate layers to quantify how much a particular layer contributes to performance on the target task \cite{shallowing}. In this work, we propose model truncation as a simple method for layer-wise pruning where the final pretrained layers after a given point are pruned off, a classification layer is appended, and this whole model is finetuned on the target task.

\section{Methods}

\subsection{Training and Evaluation Procedure}
We train chest X-ray classification models with different architectures with and without pretraining. The task of interest is to predict the probability of different pathologies from one or more chest X-rays. We use the {CheXpert} dataset consisting of 224,316 chest X-rays of 65,240 patients \cite{CheXpert} labeled for the presence or absence of 14 radiological observations. We evaluate models using the average of their AUROC metrics (AUC) on the five CheXpert-defined competition tasks (Atelectasis, Cardiomegaly, Consolidation, Edema, Pleural Effusion) as well as the No Finding task to balance clinical importance and prevalence in the validation set.

We select 16 models pretrained on ImageNet from public checkpoints implemented in PyTorch 1.4.0: DenseNet (121, 169, 201) and ResNet (18, 34, 50, 101) from \citet{pytorch}, Inception (V3, V4) and MNASNet from \citet{cadene}, and EfficientNet (B0, B1, B2, B3) and MobileNet (V2, V3) from \citet{timm}. We finetune and evaluate these architectures with and without ImageNet pretraining.

For each model, we finetune all parameters on the CheXpert training set. If a model is pretrained, inputs are normalized using mean and standard deviation learned from ImageNet. If a model is not pretrained, inputs are normalized with mean and standard deviation learned from CheXpert. We use the Adam optimizer ($\beta_1 = 0.9$,  $\beta_2 = 0.999$) with learning rate of $1\times 10^{-4}$, a batch size of 16, and a cross-entropy loss function. We train on up to four Nvidia GTX 1080 with CUDA 10.1 and Intel Xeon CPU ES-2609 running Ubuntu 16.04. For one run of an architecture, we train for three epochs and evaluate each model every 8192 gradient steps. We train each model and create a final ensemble model from the ten checkpoints, which achieved the best average CheXpert AUC across the six tasks on the validation set. We report all our results on the CheXpert test set.

We use the nonparametric bootstrap to estimate 95\% confidence intervals for each statistic. 1,000 replicates are drawn from the test set, and the statistic is calculated on each replicate. This procedure produces a distribution for each statistic, and we report the 2.5 and 97.5 percentiles as a confidence interval. Significance is assessed at the $p = 0.05$ level. 

\subsection{Truncated Architectures}
We study truncated versions of DenseNet121, MNASNet, ResNet18, and EfficientNetB0. DenseNet121 and MNASNet were chosen because we found they have the greatest efficiency (by AUC per parameters) on CheXpert of the models we profile, ResNet18 was chosen because of its popularity as a compact model for medical tasks, and EfficientNetB0 was chosen because it is the smallest current-generation model of the 16 we study. DenseNet121 contains four dense blocks separated by transition blocks before the classification layer. By pruning the final dense block and associated transition block, the model now only contains three dense blocks, yielding DenseNet121Minus1. Similarly, pruning two dense blocks and associated transition blocks yields DenseNet121Minus2, and pruning three dense blocks and associated transition blocks yields DenseNet121Minus3. For MNASNet, we remove up to the four of the final MBConv blocks to produce MNASNetMinus1 through MNASNetMinus4. For ResNet18, we remove up to the three of the final residual blocks with a similar method to produce ResNet18Minus1 through ResNet18Minus3. For EfficientNet, we remove up to two of the final MBConv6 blocks to produce EfficientNetB0Minus1 and EfficientNetB0Minus2.

After truncating a model, we append a classification block containing a global average pooling layer followed by a fully connected layer to yield outputs of the correct shape. We initialize the model with ImageNet pretrained weights, except the randomly initialized classification block, and finetune using the same training procedure as the 16 ImageNet models.

\subsection{Class Activation Maps}

We compare the class activation maps (CAMs) among a truncated DenseNet121 family to visualize their higher resolution CAMs. We generate CAMs using the Grad-CAM method \cite{gradcam}, using a weighted combination of the model's final convolutional feature maps, with weights based on the positive partial derivatives with respect to class score. This averaged map is scaled by the outputted probability so more confident predictions appear brighter. Finally, the map is upsampled to the input image resolution and overlain onto the input image, highlighting image regions that had the greatest influence on a model's decision.

\begin{figure*}[t]
\centering
\includegraphics[width=\textwidth]{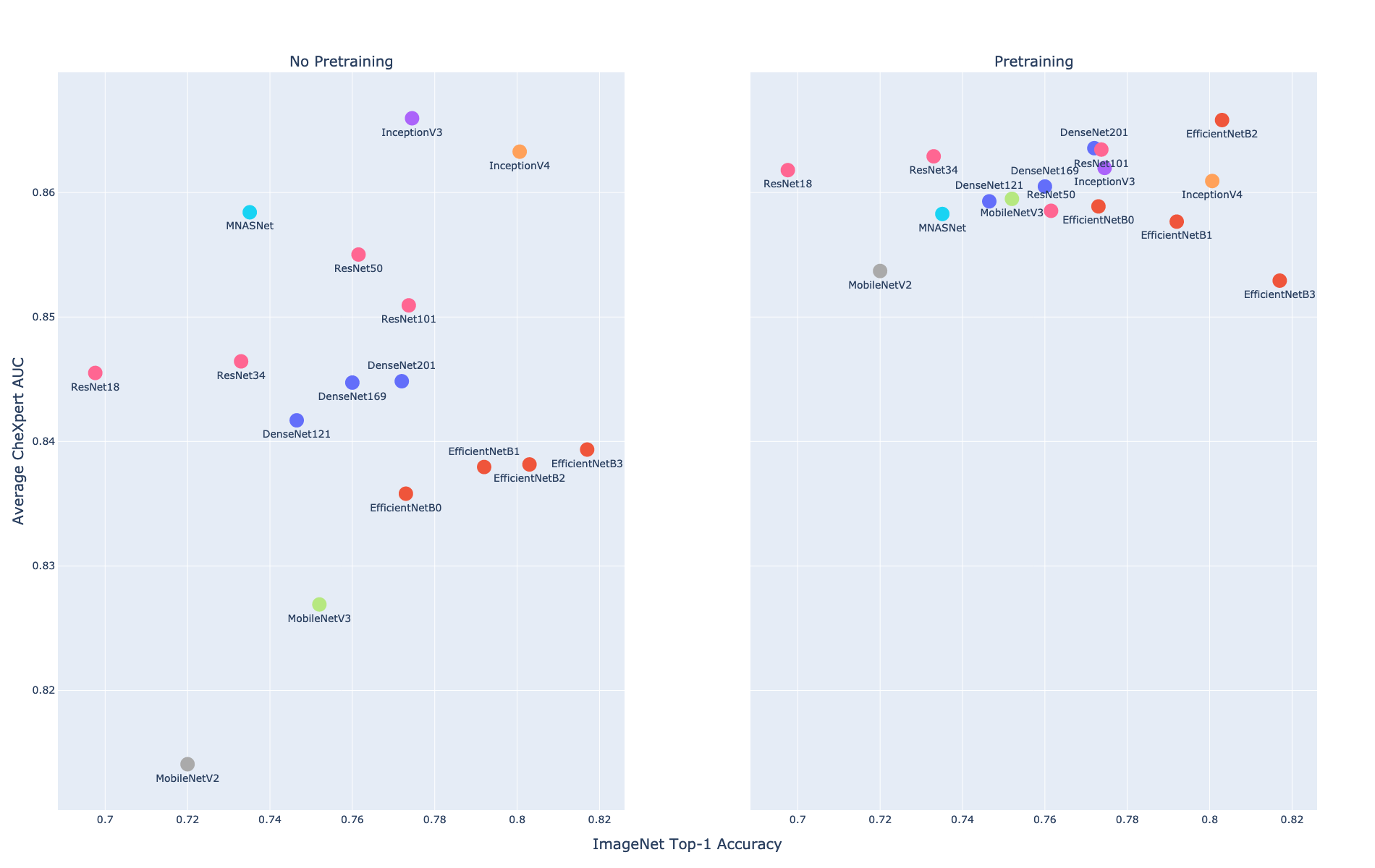} 
\caption{Average CheXpert AUC vs. ImageNet Top-1 Accuracy. The left plot shows results obtained without pretraining, while the right plot shows results with pretraining. There is no monotonic relationship between ImageNet and CheXpert performance without pretraining (Spearman $\rho = 0.08$) or with pretraining (Spearman $\rho = 0.06$).}
\Description{Two scatterplots illustrating average CheXpert AUC against the ImageNet top-1 accuracy of each model.}
\label{chexpert_vs_imagenet}
\end{figure*}

\section{Experiments}

\begin{figure*}[t]
\centering
\includegraphics[width=\textwidth]{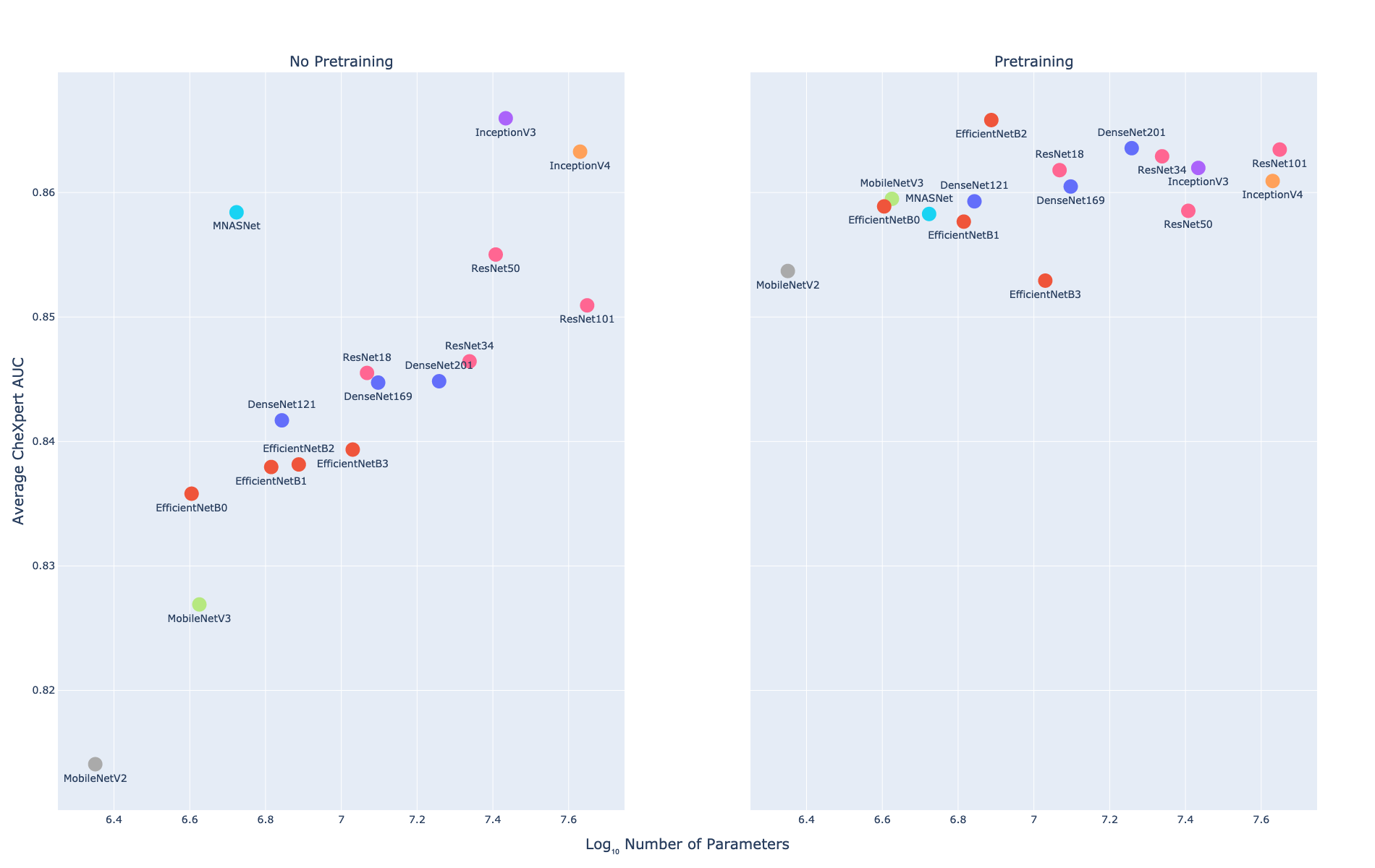} 
\caption{Average CheXpert AUC vs. Model Size. The left plot shows results obtained without pretraining, while the right plot shows results with pretraining. The logarithm of the model size has a near linear relationship with CheXpert performance when we omit pretraining (Spearman $\rho = 0.79$). However once we incorporate pretraining, the monotonic relationship is weaker (Spearman $\rho = 0.56$).}
\Description{Two scatterplots illustrating average CheXpert AUC against the base-10 logarithm of the number of parameters for each model.}
\label{chexpert_vs_params}
\end{figure*}

\subsection{ImageNet Transfer Performance}

We investigate whether higher performance on natural image classification translates to higher performance on chest X-ray classification. We display the relationship between the CheXpert AUC, with and without ImageNet pretraining, and ImageNet top-1 accuracy in Figure \ref{chexpert_vs_imagenet}

When models are trained without pretraining, we find no monotonic relationship between ImageNet top-1 accuracy and average CheXpert AUC, with Spearman $\rho = 0.082$ at $p = 0.762$. Model performance without pretraining would describe how a given architecture would perform on the target task, independent of any pretrained weights. When models are trained with pretraining, we again find no monotonic relationship between ImageNet top-1 accuracy and average CheXpert AUC with Spearman $\rho = 0.059$ at $p = 0.829$.

Overall, we find no relationship between ImageNet and CheXpert performance, so models that succeed on ImageNet do not necessarily succeed on CheXpert. These relationships between ImageNet performance and CheXpert performance are much weaker than the relationships between ImageNet performance and performance on various natural image tasks reported by \citet{imagenettransfer}. 


We compare the CheXpert performance within and across architecture families. Without pretraining, we find that ResNet101 performs only 0.005 AUC greater than ResNet18, which is well within the confidence interval of this metric (Figure \ref{chexpert_vs_imagenet}). Similarly, DenseNet201 performs 0.004 AUC greater than DenseNet121 and EfficientNetB3 performs 0.003 AUC greater than EfficientNetB0. With pretraining, we continue to find minor performance differences between the largest model and smallest model that we test in each family. We find AUC increases of 0.002 for ResNet, 0.004 for DenseNet and -0.006 for EfficientNet. Thus, increasing complexity within a model family does not yield increases in CheXpert performance as meaningful as the corresponding increases in ImageNet performance.

Without pretraining, we find that the best model studied performs significantly better than the worst model studied. Among models trained without pretraining, we find that InceptionV3 performs best with 0.866 (0.851, 0.880) AUC, while MobileNetV2 performs worst with 0.814 (0.796, 0.832) AUC. Their difference in performance is 0.052 (0.043, 0.063) AUC. InceptionV3 is also the third largest architecture studied and MobileNetV2 the smallest. We find a significant difference in the CheXpert performance of these models. This difference again hints at the importance of architecture design.

\subsection{CheXpert Performance and Efficiency}

We examine whether larger architectures perform better than smaller architectures on chest X-ray interpretation, where architecture size is measured by number of parameters. We display these relationships in Figure \ref{chexpert_vs_params} and Table \ref{all}.

\begin{table}
\centering
\begin{tabular}{@{}lrrrr@{}}
\toprule
Model          & CheXpert AUC & \#Params (M) \\ \midrule
DenseNet121	& 0.859 (0.846, 0.871)	& 6.968 \\
DenseNet169	& 0.860 (0.848, 0.873)	& 12.508 \\
DenseNet201	& 0.864 (0.850, 0.876)	& 18.120 \\
EfficientNetB0	& 0.859 (0.846, 0.871)	& 4.025 \\
EfficientNetB1	& 0.858 (0.844, 0.872)	& 6.531 \\
EfficientNetB2	& 0.866 (0.853, 0.880)	& 7.721 \\
EfficientNetB3	& 0.853 (0.837, 0.867)	& 10.718 \\
InceptionV3	& 0.862 (0.848, 0.876)	& 27.161 \\
InceptionV4	& 0.861 (0.846, 0.873)	& 42.680 \\
MNASNet	& 0.858 (0.845, 0.871)	& 5.290 \\
MobileNetV2	& 0.854 (0.839, 0.869)	& 2.242 \\
MobileNetV3	& 0.859 (0.847, 0.872)	& 4.220 \\
ResNet101	& 0.863 (0.848, 0.876)	& 44.549 \\
ResNet18	& 0.862 (0.847, 0.875)	& 11.690 \\
ResNet34	& 0.863 (0.849, 0.875)	& 21.798 \\
ResNet50	& 0.859 (0.843, 0.871)	& 25.557 \\ 
\bottomrule
\end{tabular}
\caption{CheXpert AUC (with 95\% Confidence Intervals) and Number of Parameters for 16 ImageNet-Pretrained Models.}
\label{all}
\end{table}

Without ImageNet pretraining, we find a positive monotonic relationship between the number of parameters of an architecture and CheXpert performance, with Spearman $\rho = 0.791$ significant at $p = 2.62 \times 10^{-4}$. With ImageNet pretraining, there is a weaker positive monotonic relationship between the number of parameters and average CheXpert AUC, with Spearman $\rho = 0.565$ at $p = 0.023$.


Although there exists a positive monotonic relationship between the number of parameters of an architecture and average CheXpert AUC, the Spearman $\rho$ does not highlight the increase in parameters necessary to realize marginal increases in CheXpert AUC. For example, ResNet101 is 11.1x larger than EfficientNetB0, but with only increase of 0.005 in CheXpert AUC with pretraining.

Within a model family, increasing the number of parameters does not lead to meaningful gains in CheXpert AUC. We see this relationship in all families studied without pretraining (EfficientNet, DenseNet, and ResNet) in Figure \ref{chexpert_vs_params}. For example, DenseNet201 has an AUC 0.003 greater than DenseNet121, but is 2.6x larger. EfficientNetB3 has an AUC 0.004 greater than EfficientNetB0, but is 1.9x larger. Despite the positive relationship between model size and CheXpert performance across all models, bigger does not necessarily mean better within a model family.

Since within a model family there is a weaker relationship between model size and CheXpert performance than across all models, we find that CheXpert performance is influenced more by the macro architecture design than by its size. Models within a family have similar architecture design choices but different sizes, so they perform similarly on CheXpert. We observe large discrepancies in performance between architecture families. For example DenseNet, ResNet, and Inception typically outperform EfficientNet and MobileNet architectures, regardless of their size. EfficientNet, MobileNet, and MNASNet were all generated through neural architecture search to some degree, a process that optimized for performance on ImageNet. Our findings suggest that this search could have overfit to the natural image objective to the detriment of chest X-ray tasks.


\begin{figure*}[t]
\centering
\includegraphics[width=\textwidth]{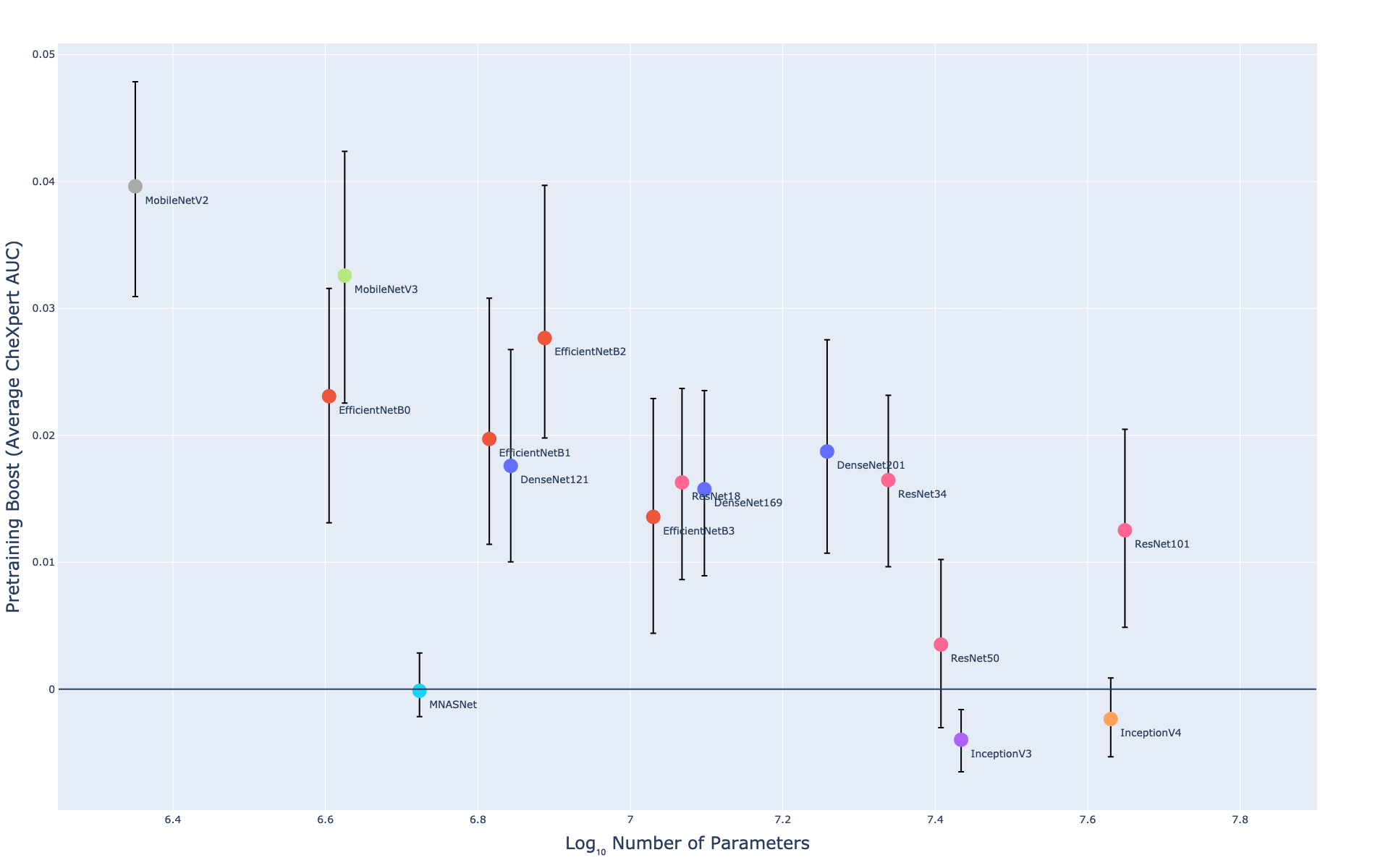} 
\caption{Pretraining Boost vs. Model Size. We define pretraining boost as the increase in the average CheXpert AUCs achieved with pretraining vs. without pretraining. Most models benefit significantly from ImageNet pretraining. Smaller models tend to benefit more than larger models (Spearman $\rho = -0.72$).}
\Description{Pretraining boost with 95-percent confidence intervals plotted against the base-10 logarithm of the number of parameters in a model.}
\label{pretrain_boost}
\end{figure*}

\subsection{ImageNet Pretraining Boost}

We study the effects of ImageNet pretraining on CheXpert performance by defining the pretraining boost as the CheXpert AUC of a model initialized with ImageNet pretraining minus the CheXpert AUC of its counterpart without pretraining. The pretraining boosts of our architectures are reported in Figure \ref{pretrain_boost}.

We find that ImageNet pretraining provides a meaningful boost for most architectures (on average 0.015 AUC). We find a Spearman $\rho = -0.718$ at $p = 0.002$ between the number of parameters of a given model and the pretraining boost. Therefore, this boost tends to be larger for smaller architectures such as EfficientNetB0 (0.023), MobileNetV2 (0.040) and MobileNetV3 (0.033) and smaller for larger architectures such as InceptionV4 ($-0.002$) and ResNet101 (0.013). Further work is required to explain this relationship. 

Within a model family, the pretraining boost also does not meaningfully increase as as model size increases. For example, DenseNet201 has a pretraining boost only 0.002 AUC greater than DenseNet121 does. This finding supports our earlier conclusion that model families perform similarly on CheXpert regardless of their size.


\subsection{Truncated Architectures}

We truncate the final blocks of DenseNet121, MNASNet, ResNet18, and EfficientNetB0 with pretrained weights and study their CheXpert performance to understand whether ImageNet models are unnecessarily large for the chest X-ray task. We express efficiency gains in terms of Times-Smaller, or the number of parameters of the original architecture divided by the number of parameters of the truncated architecture: intuitively, how many times larger the original architecture is compared to the truncated architecture. The efficiency gains and AUC changes of model truncation on DenseNet121, MNASNet, ResNet18, and EfficientNetB0 are displayed in Table \ref{truncated}.

\begin{table}
\centering
\begin{tabular}{@{}lrr@{}}
\toprule
Model                & AUC Change & Times-Smaller \\ \midrule
EfficientNetB0       & 0.00\%                & 1x                \\
EfficientNetB0Minus1 & 0.15\%                & 1.4x      \\
EfficientNetB0Minus2 & -0.45\%               & 4.7x       \\
\hline
MNASNet              & 0.00\%                & 1x                \\
MNASNetMinus1        & -0.07\%               & 2.5x        \\
MNASNetMinus2*        & -2.30\%             & 11.2x       \\
MNASNetMinus3*        & -2.51\%          & 20.0x       \\
MNASNetMinus4*        & -6.40\%               & 112.9x       \\
\hline
DenseNet121          & 0.00\%                & 1x                \\
DenseNet121Minus1    & -0.04\%               & 1.6x       \\
DenseNet121Minus2*    & -1.33\%            & 5.3x       \\
DenseNet121Minus3*    & -4.73\%             & 20.0x       \\
\hline
ResNet18             & 0.00\%                & 1x                \\
ResNet18Minus1       & 0.24\%                & 4.2x       \\
ResNet18Minus2*       & -3.70\%               & 17.1x       \\
ResNet18Minus3*       & -8.33\%            & 73.8x      
\\ \bottomrule
\end{tabular}
\caption{Efficiency Trade-Off of Truncated Models. Pretrained models can be truncated without significant decrease in CheXpert AUC. Truncated models with significantly different AUC from the base model are denoted with an asterisk.}
\label{truncated}
\end{table}


For all four model families, we find that truncating the final block leads to no significant decrease in CheXpert AUC but can save 1.4x to 4.2x the parameters. Notably, truncating the final block of ResNet18 yields a model that is not significantly different (difference -0.002 (-0.008, 0.004)) in CheXpert AUC, but is {4.2x smaller}. Truncating the final two blocks of an EfficientNetB0 yields a model that is not significantly different (difference 0.004 (-0.003, 0.009)) in CheXpert AUC, but is {4.7x smaller}. However, truncating the second block and beyond in each of MNASNet, DenseNet121, and ResNet18 yields models that have statistically significant drops in CheXpert performance.

\begin{figure}[t]
\centering
\includegraphics[width=0.95\columnwidth]{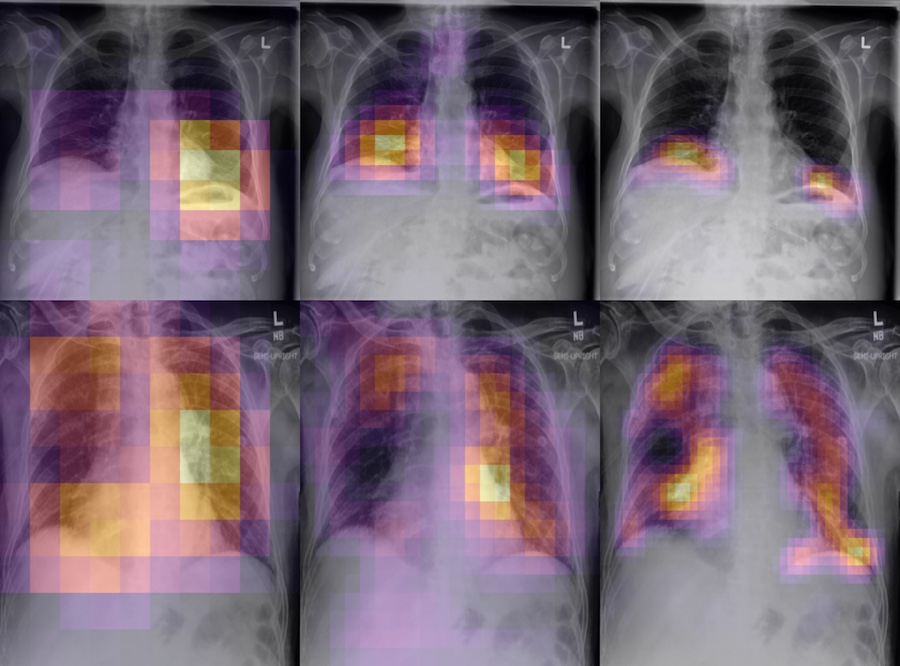} 
\caption{Comparison of Class Activation Maps Among Truncated Model Family. CAMs yielded by models, from left to right, DenseNet121, DenseNet121Minus1, and DenseNet121Minus2. Displays frontal chest X-ray demonstrating Atelectasis (top) and Edema (bottom). Further truncated models more effectively localize the Atelectasis, as well as tracing the hila and vessel branching for Edema.}
\Description{Six class activation maps of two frontal chest X-ray images showing finer detail from left to right.}
\label{cams}
\end{figure}

Model truncation effectively compresses models performant on CheXpert, making them more parameter efficient while still using pretrained weights to capture the pretraining boost. Parameter efficient models are able to lighten the computational and memory burdens for deployment to low-resource environments such as portable devices. In the clinical setting, the simplicity of our model truncation method encourages its adoption for model compression.

This finding corroborates \citet{transfusion} and \citet{bressem2020comparing}, which show simpler models can achieve performance comparable to more complex models on CheXpert. Our truncated models can use readily-available pretrained weights, which may allow these models to capture the pretraining boost and speed up training. However, we do not study the performance of these truncated models without their pretrained weights.


As an additional benefit, architectures that truncate pooling layers will also produce higher-resolution class activation maps, as shown in Figure \ref{cams}. The higher-resolution class activation maps (CAMs) may more effectively localize pathologies with little to no decrease in classification performance. In clinical settings, improved explainability through better CAMs may be useful for validating predictions and diagnosing mispredictions. As a result, clinicians may have more trust in models that provide these higher-resolution CAMs.


\section{Discussion}

In this work, we study the performance and efficiency of ImageNet architectures for chest x-ray interpretation.

\paragraph{Is ImageNet performance correlated with CheXpert?}
No. We show no statistically significant relationship between ImageNet and CheXpert performance. This finding extends \citet{imagenettransfer}---which found a significant correlation between ImageNet performance and transfer performance on typical image classification datasets---to the medical setting of chest x-ray interpretation. This difference could be attributed to unique aspects the chest X-ray interpretation task and data attributes. The chest X-ray interpretation task differs from natural image classification in that (1) disease classification may depend on abnormalities in a small number of pixels, (2) chest X-ray interpretation is a multi-task classification setup, and (3) there are far fewer classes than in many natural image classification datasets. Second, the data attributes for chest X-rays differ from natural image classification in that X-rays are greyscale and have similar spatial structures across images (always either anterior-posterior, posterior-anterior, or lateral).

\paragraph{Does model architecture matter?}
Yes. For models without pretraining, we find that the choice of architecture family may influence performance more than model size. Our findings extend \citet{transfusion} beyond the effect of ImageNet weights, since we show that architectures that succeed on ImageNet do not necessarily succeed on medical imaging tasks. A notable finding of our work is that newer architectures generated through search on ImageNet (EfficientNet, MobileNet, MNASNet) underperform older architectures (DenseNet, ResNet) on CheXpert. This finding suggests that search may have overfit to ImageNet to the detriment of medical task performance, and ImageNet may not be an appropriate benchmark for selecting architectures for medical imaging tasks. Instead, medical imaging architectures could be benchmarked on CheXpert or other large medical datasets. Architectures derived from selection and search on CheXpert and other large medical datasets may be applicable to similar medical imaging modalities including other x-ray studies, or CT scans. Thus architecture search directly on CheXpert or other large medical datasets may allow us to unlock next generation performance for medical imaging tasks.

\paragraph{Does ImageNet pretraining help?} Yes. We find that ImageNet pretraining yields a statistically significant boost in performance for chest x-ray classification. Our findings are consistent with \citet{transfusion}, who find no pretraining boost on ResNet50 and InceptionV3, but we find pretraining does boost performance for 12 out of 16 architectures. Our findings extend \citet{pretraining}---who find models without pretraining had comparable performance to models pretrained on ImageNet for object detection and image segmentation of natural images---to the medical imaging setting. Future work may investigate the relationship between network architectures and the impact of self-supervised pre-training for chest x-ray interpretation as has recently been developed by \citet{sowrirajan2020moco,azizi2021big,sriram2021covid19}.

\paragraph{Can models be smaller?} Yes. We find that by truncating final blocks of ImageNet-pretrained architectures, we can make models 3.25x more parameter-efficient on average without a statistically significant drop in performance. This method preserves the critical components of architecture design while cutting its size. This observation suggests model truncation may be a simple method to yield lighter models, using ImageNet pretrained weights to boost CheXpert performance. In the clinical setting, truncated models may provide value through improved parameter-efficiency and higher resolution CAMs. This change may enable deployment to low-resource clinical environments and further develop model trust through improved explainability.

In closing, our work contributes to the understanding of the transfer performance and parameter efficiency of ImageNet models for chest X-ray interpretation. We hope that our new experimental evidence about the relation of ImageNet to medical task performance will shed light on potential future directions for progress.

\bibliographystyle{ACM-Reference-Format}
\bibliography{sample-base}










\end{document}